\documentclass[accepted]{article}
\usepackage{natbib}
\setcitestyle{numbers,square}

\usepackage{microtype}
\usepackage{graphicx}
\usepackage{booktabs} 
\usepackage{subcaption}

\usepackage{hyperref}
\usepackage{graphicx}
\usepackage{float}
\usepackage{multirow} 
\usepackage{ragged2e} 
\usepackage{booktabs,makecell, multirow, tabularx}

\usepackage{icml2024}
\usepackage{etoolbox}
\makeatletter
\patchcmd{\@makecaption}
  {\scshape}
  {}
  {}
  {}
\makeatother

\usepackage{amsmath}
\usepackage{amssymb}
\usepackage{mathtools}
\usepackage{amsthm}

\usepackage[capitalize,noabbrev]{cleveref}

\theoremstyle{plain}
\newtheorem{theorem}{Theorem}[section]

\theoremstyle{definition}

\theoremstyle{remark}

\newcommand{\partitle}[1]{\smallskip \noindent \textbf{#1.}}
\usepackage[textsize=tiny]{todonotes}

\icmltitlerunning{Certified \textit{PEFTSmoothing}}

\begin{document}
\twocolumn[
\icmltitle{Certified \textit{PEFTSmoothing}: \\Parameter-Efficient Fine-Tuning with Randomized Smoothing}



\icmlsetsymbol{equal}{*}

\begin{icmlauthorlist}
\icmlauthor{Chengyan Fu}{equal,yyy}
\icmlauthor{Wenjie Wang}{equal,yyy}

\end{icmlauthorlist}

\icmlaffiliation{yyy}{School of Information Science and Technology, ShanghaiTech University, Shanghai, China}

\icmlcorrespondingauthor{Chengyan Fu}{fuchy2023@shanghaitech.edu.cn}
\icmlcorrespondingauthor{Wenjie Wang}{wangwj@shanghaitech.edu.cn}

\icmlkeywords{Machine Learning, ICML}

\vskip 0.3in
]




\begin{abstract}
Randomized smoothing is the primary certified robustness method for accessing the robustness of deep learning models to adversarial perturbations in the $l_2$-norm, by adding isotropic Gaussian noise to the input image and returning the majority votes over the base classifier. Theoretically, it provides a certified norm bound, ensuring predictions of adversarial examples are stable within this bound. A notable constraint limiting widespread adoption is the necessity to retrain base models entirely from scratch to attain a robust version. This is because the base model fails to learn the noise-augmented data distribution to give an accurate vote. One intuitive way to overcome this challenge is to involve a custom-trained denoiser to eliminate the noise. However, this approach is inefficient and sub-optimal. Inspired by recent large model training procedures, we explore an alternative way named \textit{PEFTSmoothing} to adapt the base model to learn the Gaussian noise-augmented data with Parameter-Efficient Fine-Tuning (PEFT) methods in both white-box and black-box settings. Extensive results demonstrate the effectiveness and efficiency of \textit{PEFTSmoothing}, which allow us to certify over 98\% accuracy for ViT on CIFAR-10, 20\% higher than SoTA denoised smoothing, and over 61\% accuracy on ImageNet which is 30\% higher than CNN-based denoiser and comparable to the Diffusion-based denoiser. 
\end{abstract}
\vspace{-1em}
\section{Introduction}
\vspace{-.5em}

\label{introduction}
\begin{figure*}[h]
\setlength{\abovecaptionskip}{0cm}
  \centering
  \includegraphics[width=.8\linewidth]{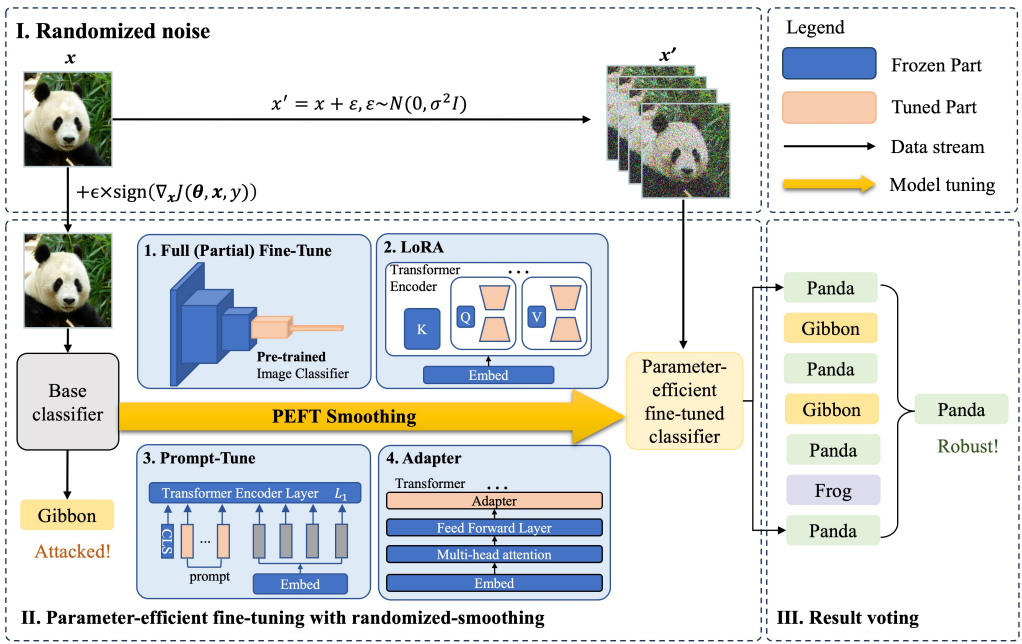}
    \vspace{-1em}
  \caption{Illustration of \textit{PEFTSmoothing} procedures. We incorporate four PEFT methods to transform a base classifier PEFTSmoothed Classifier. The final output is the majority votes of The noised-augmented inputs, which is stable within a certain norm bound.}
  \label{fig:ill}
  \vspace{-1.5em}
\end{figure*}

Certified robustness is the primary method to evaluate the robustness of deep learning systems to adversarial perturbations within specific bound \cite{pixeldp,randomized-smoothing,wordDP}. This approach offers a reliable and provable robustness guarantee that the predictions remain stable to adversarial examples within specific norm bounds. In the field of image classification, which is the fundamental task of computer vision, the SoTA certified robustness to adversarial perturbations under $l_2$-norm is randomized smoothing \cite{randomized-smoothing}. The basic idea of randomized smoothing is to convert a deterministic base classifier into a probabilistic classifier by adding isotropic Gaussian noise to the input image and returning the majority votes over the prediction of the base classifier. 

Although certified robustness holds notable theoretical strength, a significant limitation restricts its broad application: the deep learning models are required to be retrained from scratch to optimize the empirical performance as well as its robustness criterion \cite{huang2021training}. The underlying reason behind this is that the base model, initially trained on the original data distribution, fails to capture the noise-augmented data distribution. Consequently, it fails to predict the true label of the original input when subjected to the corresponding Gaussian-noised counterpart. To address this, training the base model from scratch with randomized smoothing becomes essential for better learning the noise-augmented data distribution and optimizing the certified robustness criterion. However, this requirement poses a considerable challenge, especially considering the prevalent practice of large-scale models. Not only is it impractical to train such models entirely from scratch, but it also prompts a thoughtful consideration of the trade-off between the time and computational cost involved in obtaining a robust version of a large base model.

To overcome this challenge, the most intuitive way is to incorporate a custom-trained denoiser before applying randomized smoothing \cite{denoised-smoothing}, named denoised smoothing. This denoiser processes the Gaussian-noised inputs to eliminate the noise before they reach the base classifier. The denoised inputs are expected to align with the original data distribution, enabling the model to predict the ground-truth label accurately. This process also ensures the preservation of the robustness guarantee even in the presence of Gaussian noise. Although applying a denoiser is intuitive, this approach introduces its own set of issues. First, the additional training process adds complexity to the whole architecture, demanding extra computational resources and time for training. Besides the training time, the inference time is also extremely longer than usual, especially when applying the SoTA diffusion-based denoiser architecture \cite{carlini2022certifieddiffusiondenoised-smoothing}. Second, from the previous studies, the certified accuracy of denoised smoothing is always sub-optimal, as the certified performance largely relies on the performance in the denoising procedure.

These limitations highlight the need for a more efficient and effective approach to guide large-scale models in learning noise-augmented data distributions. In this work, we explore an alternative approach—leveraging Parameter-Efficient Fine-Tuning (PEFT) \cite{xu2023parameterefficient} in conjunction with randomized smoothing. Parameter Efficient Fine-Tuning (PEFT) is an effective solution to adapting a pre-trained model to downstream tasks or datasets, by reducing the number of fine-tuning parameters and memory usage while achieving comparable performance with training from scratch \cite{xu2023parameterefficient}. Inspired by involving PEFT in recent large model training procedures, we introduced \textit{PEFTSmoothing} to acquire the model's ability to learn the underlying noise-augmented data distribution with PEFT, changing from a reactive approach (denoised smoothing) to a proactive approach (\textit{PEFTSmoothing}). The intuition behind this is that the learning capacity of the SoTA vision model such as Vision Transformer (ViT) \cite{vit} is extremely strong, it is, therefore, unnecessary to adopt an auxiliary denoiser to eliminate the noise. The intuitive nature of PEFT aligns with the inherent ability of large models to understand and adapt to noised data patterns, which is not only efficient (fewer parameters need to be tuned) but also effective (higher accuracy).

We further proposed black-boxed \textit{PEFTSmoothing} to adapt to instances where the base model is not open-sourced, for example, from a private-classification API, that white-box \textit{PEFTSmoothing} is not applicable to acquire the robust version of the base model because fine-tuning cannot be performed in this case. In addition, considering the prevailing practice in image classification, where models are commonly fine-tuned from pre-trained ones to adapt to specific downstream datasets, we further integrate the fine-tuning of \textit{PEFTSmoothing} with PEFT intended for downstream datasets adaptation. \textit{PEFTSmoothing} allows us to obtain a model that is both certifiably robust and tailored to downstream datasets, achieved through a single fine-tuning process.

In this study, we introduce \textit{PEFTSmoothing}, incorporating three prominent PEFT methods—Prompt-tuning \cite{jia2022visualvpt}, LoRA \cite{hu2021lora}, Adapter \cite{he2021towardsparalleladapter}, to guide the base model to learn the distribution of noise-augmented data. We take denoised smoothing and full(partial) fine-tuning as the baselines. Figure \ref{fig:ill} illustrates the \textit{PEFTSmoothing} procedures, including the Gaussian noise augmentation and \textit{PEFTSmoothing} to transform a base classifier to a PEFTSmoothed one and the majority vote in the inference stage, offering both theoretical robustness guarantees and empirical utility against adversarial examples. 

Our contributions can be summarized as follows:\\
    1) We reveal the insight that PEFT successfully guides large-scale models to capture the noise-augmented data distribution with modest computational and time costs. This insight explains the success of \textit{PEFTSmoothing} in converting a base model into a certifiably robust classifier (see Section \ref{sec:intuition}).\\
    2) We present \textit{PEFTSmoothing}, a certifiable method to convert large base models, such as ViT on ImageNet and CIFAR-10, into certifiably robust classifiers (see Section \ref{sec:peftsmoothing}).\\
    3) We further propose black-box \textit{PEFTSmoothing} to address scenarios where the base model cannot undergo white-box fine-tuning.\\
    4) Our experimental results demonstrate the effectiveness and efficiency of \textit{PEFTSmoothing}. In terms of accuracy, it enhances certified accuracy by over 20\% compared to denoised smoothing on CIFAR-10. On ImageNet, \textit{PEFTSmoothing} achieves comparable performance with a diffusion-based denoiser (see Section \ref{sec:experiments}). In terms of efficiency, \textit{PEFTSmoothing} reduces the training parameters by 1000 times compared to diffusion-based denoisers and 10 times compared to CNN-based denoisers. This significant reduction in training parameters indicates substantial savings in computational and time costs for obtaining a certifiably robust classifier. 

\vspace{-0.5em}
\section{Preliminaries}
\vspace{-0.5em}
Certified robustness is the major way to evaluate the robustness of deep learning models to adversarial examples. The state-of-the-art certifiable method for adversarial images in $l_2$-norm is Randomized smoothing. In this section, we first briefly review the certified guarantee of randomized smoothing. Then, we explain denoised smoothing, the practical approach to overcome the limitation of randomized smoothing which needs to train the model from scratch.
\vspace{-0.5em}
\subsection{Randomized Smoothing}
\vspace{-0.5em}
\textbf{Randomized smoothing} \cite{randomized-smoothing} converts the base classifier $\mathcal{F}$ into a smoothed classifier $\mathcal{G}$ by generating the aggregated prediction over the Gaussian noise-augmented data via majority voting. Specifically, for input \emph{x}, $\mathcal{G}$ returns the class that is most likely to be returned by the base classifier $\mathcal{F}$ under Gaussian perturbations of $x$, which can be stated as:
\begin{small}
\vspace{-1em}
\begin{equation}
\label{equ:g}
\mathcal{G}(x)=\underset{c \in \mathcal{Y}}{\arg \max } \mathbb{P}[\mathcal{F}(x+\delta)=c],
\vspace{-1em}
\end{equation}
\end{small}
where  $\delta \sim \mathcal{N}\left(0, \sigma^2 I\right)$. Under different noise scales, randomized smoothing provides a tight \emph{$l_2$} certification bound $\mathcal{R}$. Formally, the theorem can be stated as: 
\begin{theorem}
\label{theo:rs}
Given a deterministic classifier $\mathcal{F}$ and its probabilistic counterpart $\mathcal{G}$ defined in Equation \ref{equ:g}, let $\delta\sim\mathcal{N}(0,\sigma^2I)$, suppose $c_A$ is the most probable class, and $\:\underline{p_A},\overline{p_B}\in[0,1]\:$ satisfy:
\begin{small}
\vspace{-1em}
    \begin{equation}
  \mathbb{P}(\mathcal{F}(x+\varepsilon)=c_A)\geq\underline{p_A}\geq\overline{p_B}\geq\max_{c\neq c_A}\mathbb{P}(\mathcal{F}(x+\varepsilon)=c) 
  \vspace{-.5em}
\end{equation}
\end{small}
\vspace{-1em}
Then $\mathcal{G}(x+\delta)=c_A$ for all $\:\|\delta\|_2<$R$,\:$ where
\begin{small}
$$
\mathcal{R}=\frac\sigma2(\Phi^{-1}(\underline{p_A})-\Phi^{-1}(\overline{p_B}))
$$
\vspace{-1em}
\end{small}
\end{theorem}
\vspace{-2em}
$\underline{p_A}$ and $\overline{p_B}$ are the lower bound and upper bound of the top two possible classifications, and $\Phi^{-1}$ denotes the inverse of the standard Gaussian CDF. The intuition is to search for the radius that the lower bound of the highest class is still higher than the upper bound of the second highest class under certain Gaussian perturbation. Any adversarial examples within the \emph{$l_2$} ball with clean input $x$ as the center and $\mathcal{R}$ as the radius, are statistically proved to hold the same prediction results as $x$.
\vspace{-0.5em}
\subsection{Denoised Smoothing}
\vspace{-0.5em}
One limitation of randomized smoothing is that the base model needs to be trained from scratch to learn the distribution of noised input. To overcome the computation bottleneck of retraining the large-scale models, \citet{denoised-smoothing} proposed denoised smoothing to remove the noise by presenting a custom-trained denoiser $\mathcal{D}_\theta$ to any image classifier $\mathcal{F}$. The smoothed classifier $\mathcal{G}$ can be formulated as:
\begin{small}
\vspace{-0.5em}
\begin{equation}
  \mathcal{G}(x)=\underset{c \in \mathcal{Y}}{\arg \max } \mathbb{P}\left[\mathcal{F}\left(\mathcal{D}_\theta(x+\delta)\right)=c\right], 
\vspace{-1em}
\label{ds}
\end{equation}
\end{small}
where $\delta \sim \mathcal{N}\left(0, \sigma^2 I\right)$. Denoised smoothing ensures \emph{$l_p$}-robustness against adversarial examples without altering the pre-trained classifier. However, its implementation involves the training of multiple denoisers for various noise types and scales. To further improve the denoising performance and achieve higher certifiable accuracy, researchers later involved different denoiser architectures including the SoTA diffusion-based denoiser. \cite{lee2021provabledenoise,carlini2022certifieddiffusiondenoised-smoothing}.

\begin{figure}[h]
\vspace{-0.5em}
\setlength{\abovecaptionskip}{-0.2cm}
  \centering
  \includegraphics[width=.8\linewidth]{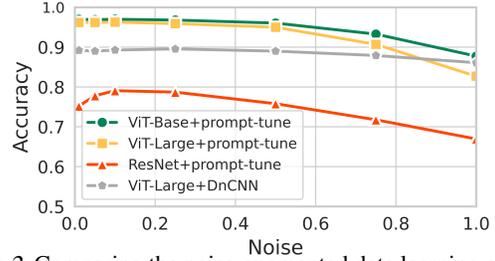}
  \vspace{-2em}
  \caption{Comparing the noise-augmented data learning capacity of Prompt-tuning and denoiser}
  \label{fig:noise}
\vspace{-1em}
\end{figure}

\vspace{-0.5em}
\section{Intuition: PEFT Guides Large Vision Models to Learn Noise-Augmented Data Distribution}
\label{sec:intuition}
\vspace{-0.5em}
In this section, we discuss the intuition behind \textit{PEFTSmoothing}, using Parameter-Efficient Fine-Tuning to achieve a certifiably robust classifier. First, we will explain why randomized smoothing empirically fails without training from scratch. Second, we will briefly review the state-of-the-art PEFT methods. Third, we will demonstrate that PEFT methods can more successfully guide large vision models to learn noise-augmented data distribution compared to eliminating the noise with denoisers. 

Theoretically, Theorem \ref{theo:rs} of randomized smoothing holds regardless of how the base classifier is trained. However, the accuracy largely relies on the majority votes on how the model classifies the Gaussian noise-augmented data. Therefore, how to improve the model's prediction ability on noise-augmented data is the key component. 

Denoised smoothing involves an auxiliary custom-trained denoiser to reactively remove the noise, allowing the model to predict the denoised input, which is close to the original data distribution. However, we assume that large-scale models such as Vision Transformer (ViT), acquire such potential to capture complex patterns and information, making them well-suited for learning intricate data distributions, including noise-augmented data. 

Parameter-Efficient Fine-Tuning, originally developed in the natural language processing (NLP) domain, aims to improve the performance of pre-trained language models on specific downstream tasks. Typical PEFT methods such as prompt-tuning leverage the existing architecture of large models, fixing its parameters and only training a small subset of parameters (soft prompt) to adapt to the downstream task or datasets. 
The intuitive nature of PEFT aligns with the inherent ability of large models to understand and adapt to diverse data patterns. Inspired by this, we propose \textit{PEFTSmoothing}, utilizing PEFT methods to guide the model to adjust its parameters more efficiently and effectively to the noise-augmented data distribution compared to the potentially sub-optimal process of training an additional denoising module. 

To substantiate our hypothesis that PEFT can more effectively guide large-scale models such as ViT to learn the noise-augmented data distribution, we compare the prediction accuracy of prompt-tuning against a DnCNN-based denoiser on noise-augmented inputs with varying Gaussian noise scales. As shown in Figure \ref{fig:noise}, for both ViT-Large and ViT-Base, prompt-tuning with noise-augmented data (yellow and green lines) consistently outperforms ViT with an auxiliary denoiser (grey line) across different noise scales.  These results indicate that for large-scale models, PEFT can better guide the model to learn the noise-augmented data distribution. Furthermore, upon comparing the accuracy trends of prompt-tuned ResNet (represented by the red line) with those of ViT (indicated by the yellow and green lines), it becomes evident that as model complexity increases, prompt-tuning excels in guiding the model to effectively learn the noise-augmented data distribution, resulting in higher accuracy.

\vspace{-.5em}
\section{\textit{PEFTSmoothing}} \label{sec:peftsmoothing}
\vspace{-0.5em}
Inspired by the above-mentioned idea that PEFT can guide the large vision models to learn noise-augmented data distribution, we propose \textit{PEFTSmoothing}, converting a base model to a robust version by fine-tuning it with Gaussian noise-augmented data and still holds the certifiable guarantee of randomized smoothing mentioned in \ref{theo:rs}. 

In this section, we first describe the white-box \textit{PEFTSmoothing} including fine-tuning on the noise-augmented data with Prompt-tuning \cite{jia2022visualvpt}, LoRA \cite{hu2021lora} and Adapter \cite{he2021towardsparalleladapter, houlsby2019parameternlpseriesadapter}. In the end, we also propose the black-box \textit{PEFTSmoothing}, considering the more general cases where the user wants to have a robust version of a larger model without access to the private classification APIs.


Given a base-classifier $\mathcal{F}$, a dataset with image $x$ and its corresponding correct label \emph{y}, we follow the standard fine-tuning method to optimize the prediction of the Gaussian perturbed input $x+\delta$ over ground-truth label \emph{y}, where $\delta \sim \mathcal{N}\left(0, \sigma^2 I\right)$. The optimization process can be stated as follows:

\vspace{-1.5em}
\begin{equation}
\underset{\theta} {\mathrm{argmin}} \quad  \mathcal{C}\mathcal{E}Loss(y, \mathcal{F}_{\theta}(x+\delta))
\label{loss}
\end{equation}
where $\mathcal{C}\mathcal{E}Loss$ refers to the standard cross entropy loss and $\mathcal{F}_{\theta}(\cdot)$ returns the probability vectors over the labels. 

We develop \textit{PEFTSmoothing} with three state-of-the-art PEFT methods including Prompt-tuning, LoRA and Adapter, as well as the full fine-tuning, which is illustrated in Figure \ref{fig:ill}. For all four fine-tuning methods, the blue blocks indicated the frozen layers which will not be optimized or updated during training while the light red blocks refer to the part that will be updated. 

\partitle{Prompt-tuning}
Prompt-tuning is a popular PEFT method that adds a trainable prefix before the embedded inputs. Similarly, in \textit{PEFTSmoothing}, soft prompts are inserted into every transformer layer's input space. Specifically, the input of the $i$th transformer layer can be stated as $[[CLS], \mathbf{P}_{i-1}, \mathbf{E}_{i-1}]$, where $\mathbf{P}_{i-1}$ refers to the trainable soft prompt of $i_{th}$ layer and $\mathbf{E}_{i-1}$ indicates the embedded input image of $i_{th}$ layer.

\partitle{LoRA}
LoRA is another typical PEFT method that freezes the pre-trained model weights and injects trainable rank decomposition matrices into each layer of the Transformer's self-attention layer, greatly reducing the number of trainable parameters for downstream tasks. In \textit{PEFTSmoothing}, LoRA weights are added to each query and value projection matrices of a ViT's self-attention layer. During fine-tuning, we freeze all the parameters of the original model and constrain the update of the layer by representing them with a low-rank decomposition. A forward path $h=\mathcal{W}_0 x$ can be modified in this way as:
\vspace{-0.5em}
\begin{equation}
    h=\mathcal{W}_0 x+\triangle \mathcal{W} x=\mathcal{W}_0 x+\mathcal{B}\mathcal{A}x
\vspace{-0.5em}
\end{equation}
where $x$ and $h$ denote the input and output features of each layer, $\mathcal{W}_0 \in \mathbb{R}^{d \times k}$ indicates the original weights of base model $\mathcal{F}$ while $\triangle W$ denotes the weight change which is composed of the inserted low-rank matrices $\mathcal{B} \in \mathbb{R}^{d \times r}$ and $\mathcal{A} \in \mathbb{R}^{r \times k}$.

\partitle{Adapter}
The series adapter is applied directly after the attention and feed feedforward layer of each sub-layer. In \textit{PEFTSmoothing}, we add small adapters into each transformer layer. The architecture of an adapter includes a down-projection with $\mathcal{W}_{Down}$ to project the input to a lower-dimensional space specified by bottleneck dimension $r$, followed by a nonlinear activation function $f(\cdot)$, and an up-projection with $\mathcal{W}_{Up}$, which can be stated as:
\vspace{-0.8em}
$${h = \mathcal{W}_0x+f(\mathcal{W}_0x\mathcal{W}_\mathrm{down}}){\mathcal{W}_\mathrm{up}}$$
where similarly, $x$ and $h$ denote the input and output features of each layer, $\mathcal{W}_0$ indicates the original weights of base model.

\partitle{Black-box \textit{PEFTSmoothing}}
To adapt \textit{PEFTSmoothing} to more general cases where the base model is a black-box one that common white-box PEFT methods are not applicable, we also propose black-box \textit{PEFTSmoothing} utilizing black-box prompt-tuning \cite{bahng2022visualvp,oh2023blackvip}. The advantage of adapting prompt-tuning to the black-box version is that black-box prompt-tuning is independent of the base model parameters and does not involve the modification of the main model architecture. The basic idea of black-box prompt-tuning is to generate custom pixel-style prompts by a learnable autoencoder \cite{ballard1987modularae} via approximating the high-dimensional gradients with Simultaneous Perturbation Stochastic Approximation (SPSA) \cite{spall1992multivariateSPSA1,spall1997oneSPSA2} instead of directly calculating the gradients in the white-box setting. More specifically, we first build an autoencoder-based Coordinator, consisting of a frozen encoder $f(\cdot)$ and a lightweight learnable decoder $g_{\phi_d}(\cdot)$ with parameter $\phi_d$. The pixel-style prompts are generated by the Coordinator and added to the image $x$. To have a prompt-injected data with the Gaussian noise, we have:
\vspace{-0.5em}
$$\tilde{x}=clip(x+\epsilon h_{\boldsymbol{\phi}}(x))+\delta, \quad h_\phi(x)=g_{\phi_d}(z_x,\phi_t),$$
where $\delta \sim \mathcal{N}\left(0, \sigma^2 I\right)$ and $\phi_{t}$ is a task-specific prompt trigger vector that is jointly optimized with decoder parameter $\phi_d$. Here $z_{x}=f(x)$ is the output of the encoder and $\epsilon\in[0,1]$ is the hyperparameter to control the power of the prompt.
\vspace{-0.5em}
\section{Experiments}\label{sec:experiments}
\vspace{-0.5em}
In this section, we first elaborate on the experiment configurations. Second, we evaluate \textit{PEFTSmoothing} on the Vision Transformer (ViT) for CIFAR-10 and ImageNet in terms of certified accuracy and computation costs, compared with baseline methods including basic randomized smoothing and denoised smoothing. Third, we conduct ablation studies on Prompt-tuning and LoRA to demonstrate how the selection of hyper-parameters influences the certified results. Then, we further evaluate the black-box \textit{PEFTSmoothing} from the perspective of certified accuracy. Last, we present the possibility of integrating the fine-tuning of \textit{PEFTSmoothing} with PEFT intended for downstream dataset adaptation. 
\begin{figure*}[h]
    \centering
    \begin{subfigure}{.4\textwidth}
          \includegraphics[width=\textwidth]{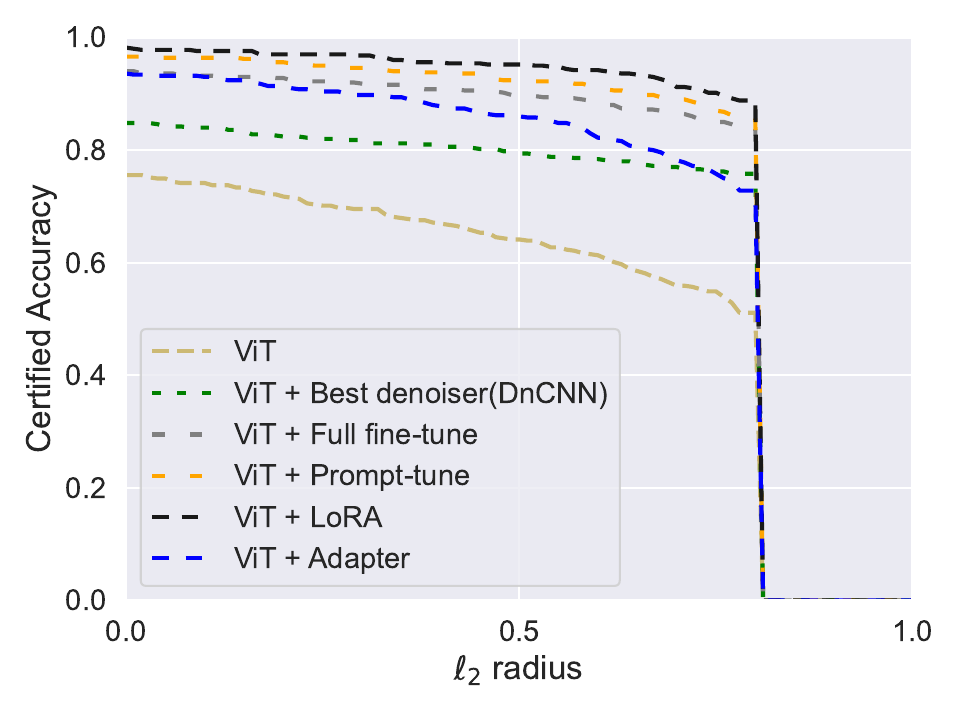}
          \vspace{-2em}
          \caption{CIFAR-10 $\sigma=0.25$}
          \label{methodsscifar0.25}
        \end{subfigure}
        \begin{subfigure}{.4\textwidth}
          \includegraphics[width=\textwidth]{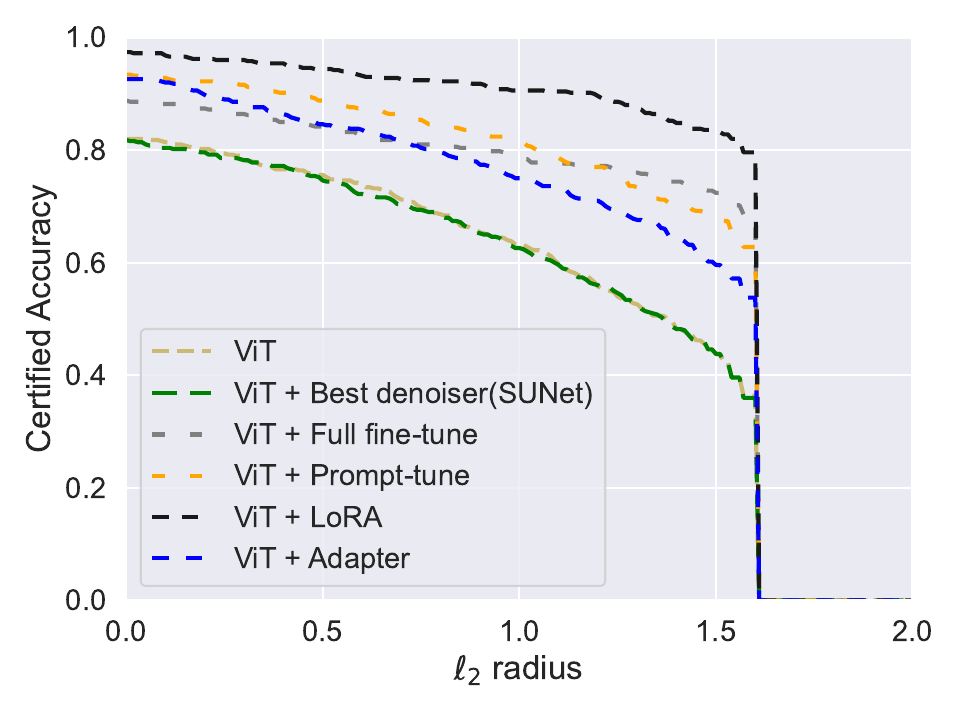}
          \vspace{-2em}
          \caption{CIFAR-10 $\sigma=0.5$}
          \label{methodsscifar0.5}
        \end{subfigure}
        \begin{subfigure}{.4\textwidth}
          \includegraphics[width=\textwidth]{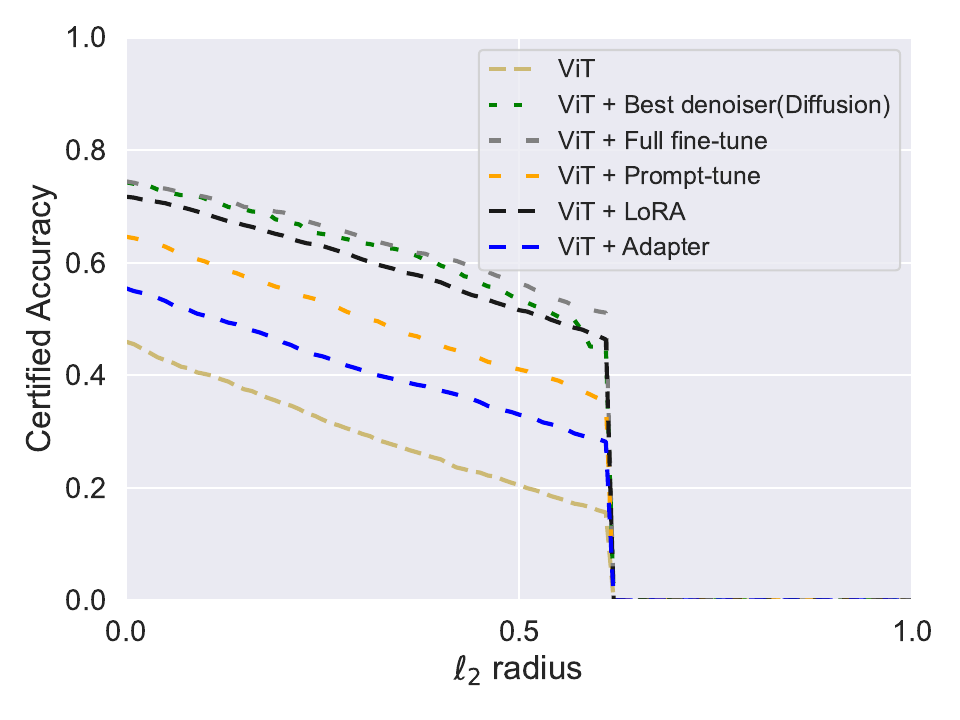}
          \vspace{-2em}
          \caption{ImageNet $\sigma=0.25$}
          \label{methodssimagenet0.25}
        \end{subfigure}
        \begin{subfigure}{.4\textwidth}
          \includegraphics[width=\textwidth]{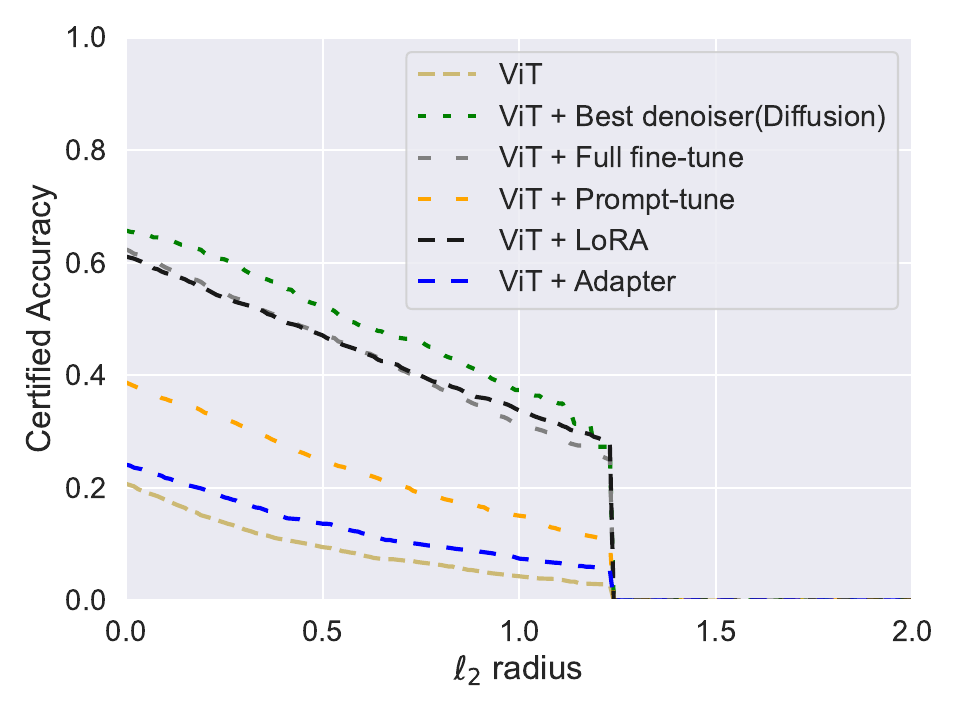}
          \vspace{-2em}
          \caption{ImageNet $\sigma=0.5$}
          \label{methodssimagenet0.5}
        \end{subfigure}
    \vspace{-1em}
    \caption{Certified accuracy comparison on \textit{PEFTSmoothing} and denoised smoothing }
    \label{cifarmethods}
\vspace{-2em}
\end{figure*}

\begin{table*}[t]
\vskip 0.15in
\begin{center}
\begin{small}
\begin{sc}
\resizebox{2\columnwidth}{!}{
\begin{tabular}{clcccccc}
\toprule
&&\multicolumn{4}{c|}{CIFAR-10}&\multicolumn{2}{c}{ImageNet}\\
Category & Method & $\sigma=0.25$ & $\sigma=0.50$ & $\sigma=0.75$ & $\sigma=1.00$ & $\sigma=0.50$ & $\sigma=1.00$\\
\midrule
\multirow{10}{*}{RS} & PixelDP \cite{pixeldp}    & $22.0^{(71.0)}$ & $2.0^{(44.0)}$ & \--{} & \--{} & $16.0^{(33.0)}$ & \--{}\\
&RS \cite{randomized-smoothing} & $61.0^{(75.0)}$& $43.0^{(75.0)}$& $32.0^{(65.0)}$ & $22.0^{(66.0)}$& $49.0^{(67.0)}$ & $37.0^{(57.0)}$\\
&SmoothAdv \cite{salman2019provablysmoothadv}    & $67.4^{(75.6)}$& $57.6^{(75.6)}$& $47.8^{(74.7)}$ & $38.3^{(57.4)}$ & $56.0^{(65.0)}$ & $43.0^{(54.0)}$\\
&SmoothAdv \cite{salman2019provablysmoothadv}    & $74.9^{(74.3)}$& $63.4^{(80.1)}$&$ 51.9^{80.1)}$ & $39.6^{(62.2)}$ & \--{} & \--{}\\
&Consistency \cite{jeong2020consistency}     & $68.8^{(77.8)}$& $58.1^{(75.8)}$& $48.5^{72.9)}$ & $37.8^{(52.3)}$ & $50.0^{(55.0)}$ & $44.0^{(55.0)}$\\
&MACER \cite{zhai2020macer}      & $71.0^{(81.0)}$& $59.0^{(81.0)}$& $46.0^{66.0)}$ & $38.0^{(66.0)}$ & $57.0^{(68.0)}$ & $43.0^{(64.0)}$\\
&Boosting \cite{horvath2021boosting}       & $70.6^{(83.4)}$& $60.4^{(76.8)}$& $52.4^{71.6)}$ & $38.8^{(52.4)}$ & $57.0^{(65.6)}$ & $44.6^{(57.0)}$\\
&DRT \cite{yang2021certifiedensemble}       & $70.4^{(81.5)}$& $60.2^{(72.6)}$&$ 50.5^{71.9)}$ & $39.8^{(56.1)}$ & $46.8^{(52.2)}$ & $44.4^{(55.2)}$\\
&SmoothMix \cite{jeong2021smoothmix}       & $67.9^{(77.1)}$&$ 60.4^{(76.8)}$& $52.4^{71.6)} $& $38.8^{(52.4)}$ & $50.0^{(55.0)}$ & $43.0^{(55.0)}$\\
&ACES \cite{horvath2022robust}       &$ 69.0^{(79.0)}$& $57.2^{(74.2)}$& $47.0^{74.2)}$ & $37.8^{(58.6)}$ & $54.0^{(63.8)}$ & $42.2^{(57.2)}$\\
\midrule
\multirow{3}{*}{DS} & Denoised \cite{denoised-smoothing}    & $56.0^{(72.0)}$ & $41.0^{(62.0)}$ & $28.0^{(62.0)}$ & $19.0^{(44.0)}$ & $33.0^{(60.0)}$ & $14.0^{(38.0)}$\\
&Lee \cite{lee2021provabledenoise}       & 60.0& 42.0& 28.0 & 19.0 & 41.0 & 24.0\\
&Diffusion \cite{carlini2022certifieddiffusiondenoised-smoothing}       & $76.7^{(88.1)}$& $63.0^{(88.1)}$& $45.3^{(88.1)}$ & $32.1^{(77.0)}$ & $\textbf{71.1}^{(82.8)}$ & $\textbf{54.3}^{(77.1)}$\\
\midrule
\multirow{4}{*}{PEFT} & LoRA(Ours)    & $\textbf{98.2}^{(97.0)}$ & $\textbf{97.4}^{(95.4)}$ & $\textbf{95.0}^{(94.4)} $& $\textbf{94.6}^{(91.4)}$ & $61.08^{(3.9)}$ & $39.0^{(1.00)}$\\
&Prompt-tune(Ours)       & $96.6^{(\textbf{97.2})}$& $93.4^{(\textbf{97.2})}$& $93.2^{(\textbf{95.8})}$ & $87.4^{(\textbf{94.0})}$ & $38.72^{(53.9)}$ & $16.0^{(42.2)}$\\
&Full fine-tune(Ours)       & $94.0^{(94.0)}$& $88.8^{(89.0)}$& $86.6^{(84.8)}$ & $83.6^{(82.8)}$ & $62.36^{(44.1)}$ & $34.7^{(18.3)}$\\
&Adapter(Ours)       & $93.6^{(91.2)}$& $92.6^{(89.8)}$& $87.8^{(90.2)}$ & $82.8^{(87.4)}$ & $24.12^{(23.9)}$ & $11.4^{(0.080)}$\\
\bottomrule
\end{tabular}
}
\caption{CIFAR-10 certified top-1 accuracy for prior defenses of randomized smoothing, denoised smoothing, and \textit{PEFTSmoothing}. Each entry lists the certified accuracy, with the clean accuracy for that model in parentheses, using numbers taken from respective papers.}
\label{cifarresultsyable}
\end{sc}
\end{small}
\end{center}
\vskip -0.1in
\end{table*}

\begin{center}
\begin{table*}[h!]
\centering
\arraybackslash
\resizebox{.8\textwidth}{!}
{
\begin{tabular}{|c|ccc|cccc|}
\hline
\multirow{2}{*}{Method} & \multicolumn{3}{c|}{Denoised Smoothing}                              & \multicolumn{4}{c|}{\textit{PEFTSmoothing}}                                                                             \\ \cline{2-8} 
                        & \multicolumn{1}{c|}{SUNet} & \multicolumn{1}{c|}{Diffusion} & DnCNN  & \multicolumn{1}{c|}{Full Fine-Tune} & \multicolumn{1}{c|}{Adaptor} & \multicolumn{1}{c|}{Lora}   & Prompt-Tune \\ \hline
Parameters              & \multicolumn{1}{c|}{99.7M} & \multicolumn{1}{c|}{52.5M}     & 0.558M & \multicolumn{1}{c|}{86.6M}          & \multicolumn{1}{c|}{3.39M}   & \multicolumn{1}{c|}{0.081M} & 0.929M      \\ \hline
\end{tabular}
}
\vspace{-1em}
\caption{Comparison of the computational cost from the perspective of trained parameters. Base classifier is ViT-base.}
\label{tab:para}
\vspace{-1em}
\end{table*}
\end{center}
\vspace{-5em}
\subsection{Configuration}
\vspace{-.5em}
We evaluate \textit{PEFTSmoothing} results on two standard datasets, CIFAR-10 \cite{cifar10} and ImageNet2012. Form the experimental results, \textit{PEFTSmoothing} surpasses all baseline methods on the low-resolution image dataset and surpasses some of the baseline methods on high resolution one. On both datasets, We draw noisy samples to certify the robustness of base classifiers following \cite{randomized-smoothing}. All experiments of ImageNet are conducted on an A100 GPU and CIFAR-10 on an A40 GPU.

\vspace{-0.5em}
\partitle{Model configuration}  The base classifier we used to test performance of \textit{PEFTSmoothing} on CIFAR-10 is a 86.6M-parameter ViT-B/16 model \cite{vit} pre-trained on ImageNet-21k \cite{deng2009imagenet} and fine-tuned to CIFAR-10. For ImageNet,  we also used the same pre-trained ViT-B/16 model and but fine-tuned on ImageNet2012 by Google \cite{vit}. 

\vspace{-0.5em}
\partitle{Baselines} 
For baseline comparison, we mainly compared the performance and the computation costs of \textit{PEFTSmoothing} with the SoTA denoised smoothing, namely DnCNN-based \cite{zhang2017beyonddncnn} and diffusion-based denoiser \cite{nichol2021improveddiffusiondenoiser}. Besides, we also use a state of art SUNet-structured (swin transformer + UNet) denoiser \cite{fan2022sunet} which is used in medical image denoising as a new baseline denoiser. 



\vspace{-0.5em}
\partitle{\textit{PEFTSmoothing} configuration} For Prompt-tuning, we add soft prompts as prefixes before the input embedding of an image with a length of 100. For LoRA, we add trainable rank decomposition matrices into each layer of the Transformer architecture with the rank of 2. For the Adapter, we insert new MLP modules with residual connections inside Transformer layers. The hyper-parameter selection and ablation study will be discussed in the sections below. For the black-box \textit{PEFTSmoothing}, we adopt a CLIP ViT-B/16 \cite{radford2021learningclip} as the pre-trained model and an ImageNet pre-trained $\textit{vit-mae-base}$ as the frozen encoder of Coordinator which is introduced above in Section \ref{sec:peftsmoothing}.

\vspace{-0.5em}
\partitle{Evaluation metrics}
Certified accuracy is the standard metric to evaluate the robustness of the defense methods. Certified accuracy denotes the fraction of the clean testing set on which the predictions are both correct and satisfy the certification criteria (see Theorem \ref{theo:rs}). 
\vspace{-1em}
\subsection{Certified Accuracy of \textit{PEFTSmoothing}}
\vspace{-0.5em}
Figure \ref{cifarmethods} shows the comparison of certified accuracy between \textit{PEFTSmoothing} and denoised smoothing. The upper two figures and the lower two figures represent the certified results on CIFAR-10 and ImageNet respectively under different Gaussian noise scales $\sigma$. Note that for denoiser, we only demonstrate the best denoised smoothing results, DnCNN-based and SUNet-based on CIFAR-10 and Diffusion-based denoiser on ImageNet.

As shown in Figure \ref{methodsscifar0.25} and \ref{methodsscifar0.5}, \textit{PEFTSmoothing} outperforms all the SoTA denoised smoothing approaches on CIFAR-10. \textit{PEFTSmoothing} has not only the highest certified accuracy when \emph{$l_2$} radius = 0, but also decreases the most gradually when \emph{$l_2$} radius increases. Surprisingly, LoRA and Prompt-tune even outperform full fine-tuning on both settings, while training significantly fewer total model parameters (LoRA: 0.081M, Prompt-tune: 0.929M, Adapter: 3.387M \emph{vs} Full: 86.6M). Thus even though storage is not a concern, \textit{PEFTSmoothing} is a promising approach for processing a base classifier for randomized-smoothing.

Nevertheless, the experiments on ImageNet (Figure \ref{methodssimagenet0.25} and Figure \ref{methodssimagenet0.5}) experience a different trend where the denoised smoothing with Diffusion-based denoiser has a slightly higher certified accuracy than \textit{PEFTSmoothing} with LoRA. This is mainly because of the powerful denoising ability inherent in the intricate diffusion architecture, especially for high resolution images. However, it's noteworthy that while this complex architecture enhances the denoising process, it simultaneously results in significantly prolonged inference times for individual examples, given that all noise-augmented data must traverse the Diffusion-based denoiser.

To conduct a more thorough comparison with basic randomized smoothing methods as well as the denoised smoothing, in Table \ref{cifarresultsyable}, we report the top-1 certified accuracy achieved by \textit{PEFTSmoothing} and other baseline methods for different noise magnitudes on two datasets. On CIFAR-10, \textit{PEFTSmoothing} outperforms all randomized smoothing and denoised smoothing methods at all noise magnitudes greatly. All four \textit{PEFTSmoothing} methods can achieve over 80\% top-1 certified accuracy at high $\sigma$ distortions ($\sigma >= 0.5$) and can achieve over 90\% at low $\sigma$ distortions while the state of art denoised smoothing and randomized smoothing methods can achieve at most around 75\% at $\sigma = 0.25$. Furthermore, among all four \textit{PEFTSmoothing} methods, LoRA performs the best top-1 accuracy which can maintain over 94\% over all $\sigma$ and prompt-tune performs the best accuracy on clean dataset but top-1 certified accuracy slightly worse than LoRA. The results on ImageNet reveals the same trend as Figure \ref{methodssimagenet0.25} and Figure \ref{methodssimagenet0.5}, that denoised smoothing with Diffusion-based model has the best certified accuracy, around 10\% higher than the best \textit{PEFTSmoothing} with LoRA, due to the same reason that Diffusion-based denoiser has high denoising ability on high-resolution inputs. 

\begin{figure}[H]
  \vspace{-1em}
\setlength{\abovecaptionskip}{1em}
  \centering
  \includegraphics[width=.7\linewidth]{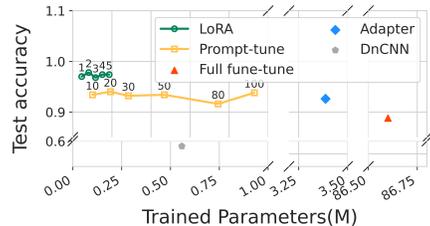}
  \vspace{-1.8em}
  \caption{Test Accuracy vs. Size of Trained Parameters}
  \label{fig:para}
\vspace{-1.5em}
\end{figure}

In terms of efficiency, we demonstrate the size of training parameters in Table\ref{tab:para}. \textit{PEFTSmoothing} with LoRA, Adapter, and Prompt-tuning reduces the training parameters by 1000 times compared to Diffusion-based denoisers and 10 times compared to DnCNN-based denoisers. This significant reduction in training parameters indicates substantial savings in computational and time costs for obtaining a certifiably robust classifier. We also demonstrate the size of trained parameters vs. achieved certified accuracy in Figure \ref{fig:para}, comparing \textit{PEFTSmoothing} with the denoised smoothing under different ablation settings. Ideally, we want to achieve high certified accuracy with low size of trained parameters (upper left part of the graph), which is dominated by  \textit{PEFTSmoothing} with LoRA and Prompt-tuning.

\begin{figure}[h]
  \vspace{-0.8em}
\setlength{\abovecaptionskip}{0cm}
  \centering
  \includegraphics[width=.7\linewidth]{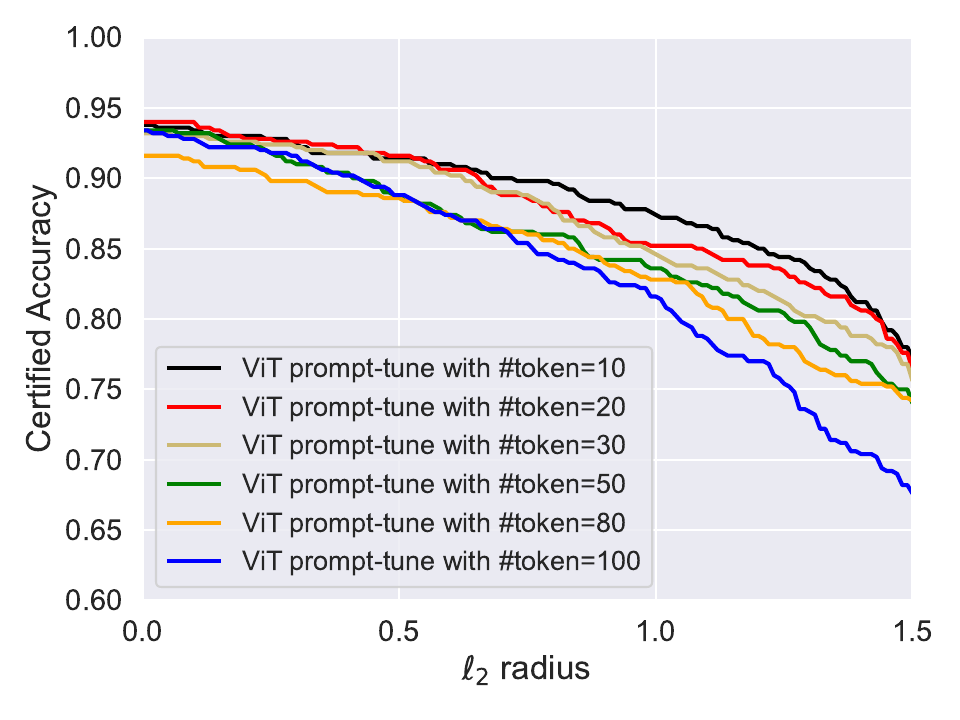}
  \vspace{-2em}
  \caption{Certified Accuracy of Prompt-tuning in \textit{PEFTSmoothing} with Different Prompt Lengths. $\sigma=0.5$}
  \label{ablationprompt}
\vspace{-0.5em}
\end{figure}

\vspace{-1em}
\subsection{CIFAR-10 Ablation Studies}
\begin{figure}[H]
  \vspace{-1em}
\setlength{\abovecaptionskip}{1em}
  \centering
  \includegraphics[width=.7\linewidth]{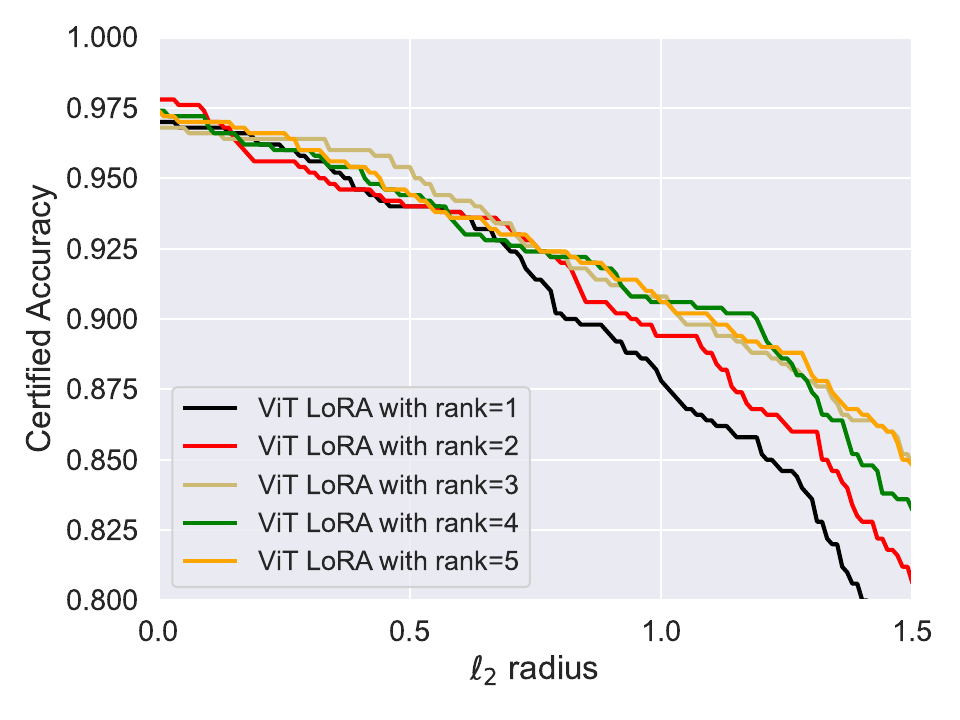}
  \vspace{-1.8em}
  \caption{Certified Accuracy of LoRA in \textit{PEFTSmoothing} with Different Ranks. $\sigma=0.5$}
  \label{ablationrank}
\vspace{-1em}
\end{figure}

\vspace{-0.5em}
In this section, we present the results on the ablation studies of \textit{PEFTSmoothing} with prompt-tuning and LoRA in terms of Prompt Lengths and LoRA Ranks respectively, with the same setup in figure \ref{cifarmethods}. Figure \ref{ablationprompt} shows the certified accuracy of \textit{PEFTSmoothing} with Prompt-tuning with different prompt lengths and Figure \ref{ablationrank} shows part of the certified accuracy of \textit{PEFTSmoothing} with LoRA with different LoRA ranks. The certified accuracy doesn't change much among different prompt lengths and different ranks. Notably, even with as few as only 10 prompts or $rank=1$,  \textit{PEFTSmoothing} still achieves high certified accuracy and remains competitive or even better compared to full fine-tuning and other certified defense methods, indicating the ability to guide the model to learn the noised inputs. 

\vspace{-1em}
\subsection{Black-box \textit{PEFTSmoothing}}

\begin{figure}[H]
\setlength{\abovecaptionskip}{0cm}
  \centering
  \includegraphics[width=.65\linewidth]{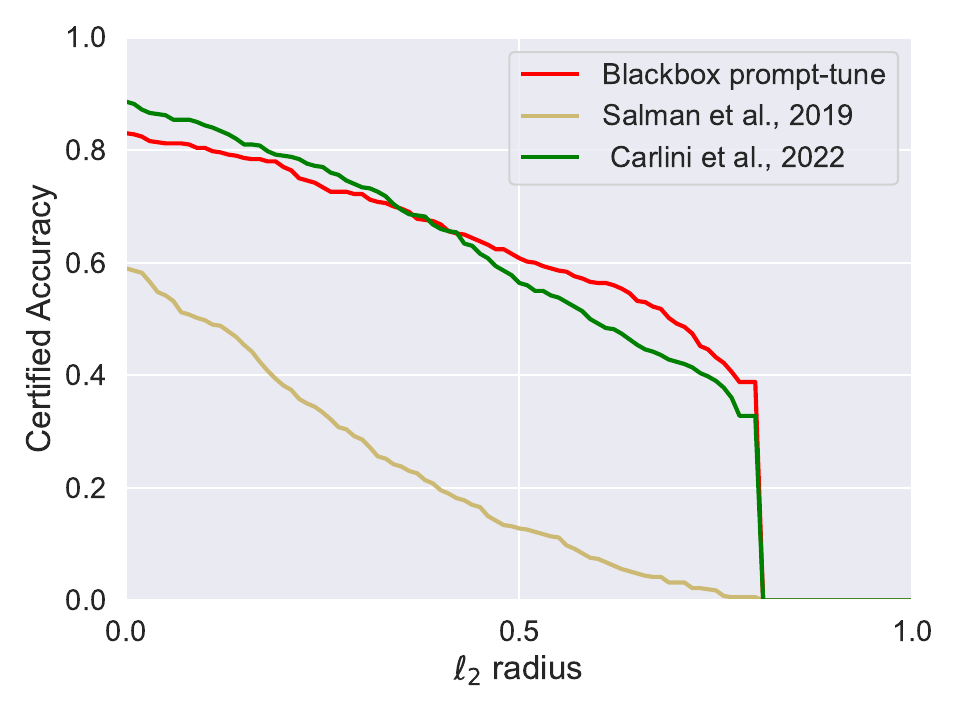}
    \vspace{-2em}
  \caption{Certified Accuracy of Black-box \textit{PEFTSmoothing}}
  \label{blackbox}
    \vspace{-1em}
\end{figure}

\vspace{-0.5em}
To demonstrate the effectiveness of black-box \textit{PEFTSmoothing}, we compare the certified accuracy under Gaussian noise scale $\sigma$ equals 0.25 on CIFAR-10 with DnCNN-based denoiser \cite{denoised-smoothing} and Diffusion-based denoiser \cite{carlini2022certifieddiffusiondenoised-smoothing} in Figure \ref{blackbox}, which are both certified defenses under black-box settings. As illustrated in the figure, black-box \textit{PEFTSmoothing} greatly outperforms DnCNN-based denoised smoothing \cite{denoised-smoothing} with top-1 certified accuracy of 0.83 against 0.6. Meanwhile, black-box \textit{PEFTSmoothing} can achieve similar performance to diffusion-based denoised smoothing \cite{carlini2022certifieddiffusiondenoised-smoothing} since our method has better-certified accuracy at large radius ($\ell_{2} radius > 0.4$) and nearly match it at small radius. 
\vspace{-1em}
\vspace{-1em}
\subsection{Two Birds One Stone}
\begin{figure}[H]
\setlength{\abovecaptionskip}{0cm}
  \centering
  \includegraphics[width=.65\linewidth]{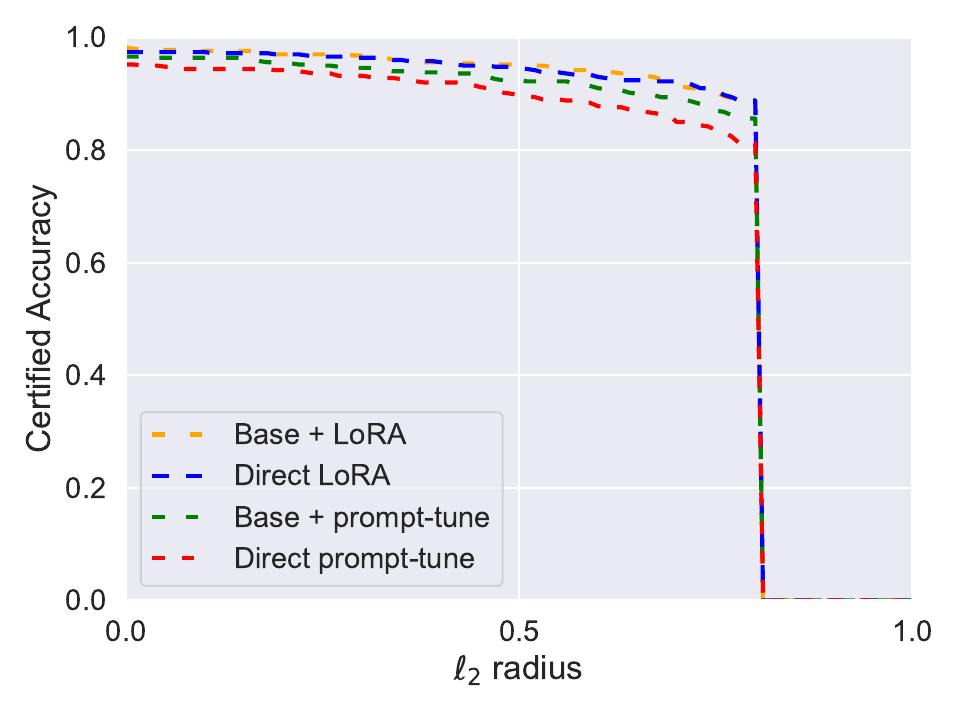}
    \vspace{-2em}
  \caption{Integrating the fine-tuning of \textit{PEFTSmoothing} with PEFT intended for downstream dataset adaptation}
  \label{direct}
    \vspace{-1em}
\end{figure}

\vspace{-0.5em}
In this section, we present the possibility of integrating the fine-tuning of \textit{PEFTSmoothing} with PEFT intended for downstream dataset adaptation. As shown in Figure \ref{direct}, one round of PEFT to achieve both \textit{PEFTSmoothing} and downstream dataset (blue and red lines) can achieve comparable performance with fine-tuning to the datasets and \textit{PEFTSmoothing} sequentially (yellow and green lines). Considering the prevailing practice in image classification, where models are commonly fine-tuned from pre-trained ones to adapt to the specific downstream datasets, these results indicate the promising direction of achieving a certifiable robust version of deep learning systems together with downstream adaptations for free. 
\vspace{-1em}

\section{Related Work}
\vspace{-1em}
\partitle{Adversarial attack} \citet{goodfellow} first brought attention to adversarial attacks by introducing the concept of adversarial examples and demonstrating their ability to deceive machine learning models. Later \cite{carlini2017} found stronger attack methods by introducing optimization-based methods for crafting adversarial examples. 

\vspace{-1em}
\partitle{Empirical defense} The most successful empirical defense is adversarial training \cite{goodfellowadtrain,kannan2018adversarial}. During the training phase, the model is exposed to adversarial examples. Researchers later investigated the effectiveness of adversarial training across different models and proposed ensemble adversarial training \cite{tramer2017adensemble}. However, many heuristic defenses were later easily broken by some stronger adversarial attack methods \cite{carlini2017adversarial, athalye2018adobfuscated, athalye2018adrobustness, uesato2018adversarial}. 

\vspace{-1em}
\partitle{Certified robustness} Certified defense refers to providing provable guarantees on the robustness and reliability of models against adversarial attacks. PixelDP \cite{pixeldp} can scales to large networks and datasets which is based on a novel connection between robustness against adversarial examples and differential privacy. Randomized smoothing \cite{randomized-smoothing} is then developed to create a robustness guarantee by modeling the inherent uncertainty in the decision boundaries of the model through incorporating random noise into the input data. Researchers later found different ways to improve the existing randomized smoothing algorithm. \cite{salman2019provablysmoothadv} introduced adversarial training into the training stage. \cite{jeong2020consistency} tried to regularize the prediction consistency over noise. \cite{horvath2021boosting,yang2021certifiedensemble} studied better robustness of ensemble models over single model. A simple training scheme called SmoothMix \cite{jeong2021smoothmix} is then proposed to control the robustness of smoothed classifiers via self-mixup. \citet{horvath2022robust} studied the balance between robustness and accuracy by proposing a compositional architecture. Researchers also considered adding a denoiser before the base classifier to reduce the influence of Gaussian noise on final prediction \cite{denoised-smoothing,carlini2022certifieddiffusiondenoised-smoothing,lee2021provabledenoise}.

\vspace{-1em}
\partitle{Fine-tune} Fine-tuning is a widely employed technique in machine learning, involving taking a pre-trained model and further training it on a specific task or dataset to enhance its performance. However, for nowadays large-scale models, researchers found it time consuming and computational-consuming to tune the whole model \cite{houlsby2019parameternlpseriesadapter} as they propose Parameter-efficient fine-tuning to emphasize the optimization of model parameters with a focus on resource efficiency, including Prompt-tuning \cite{jia2022visualvpt}, LoRA \cite{hu2021lora}, Adapter \cite{he2021towardsparalleladapter}. 

\vspace{-1em}                                                                                           
\section{Conclusion}
\vspace{-1em}
In this paper, we present \textit{PEFTSmoothing} to proactively adapt the base model to learn the Gaussian noise-augmented data distribution with Parameter-Efficient Fine-Tuning methods. PEFTSmoothed model can achieve high certified accuracy when applying randomized smoothing procedures. We experimented \textit{PEFTSmoothing} with different PEFT strategies and compared them with basic randomized smoothing and denoised smoothing. Experimental results indicate that \textit{PEFTSmoothing} greatly outperforms the existing certified defense methods on CIFAR-10 and ImageNet while significantly decreasing the computational cost of the defense. We further explored the black-box \textit{PEFTSmoothing} and the possibility of achieving a PEFTSmoothed model along with fine-tuning to adapt to downstream datasets. Extensive experiments demonstrate the effectiveness and efficiency of \textit{PEFTSmoothing}.


\newpage
\appendix
\onecolumn

\end{document}